# Prediction of ICD Codes with Clinical BERT Embeddings and Text Augmentation with Label Balancing using MIMIC-III

Brent Biseda
University of California, Berkeley
brentbiseda@berkeley.edu

Haifeng Lin
University of California, Berkeley
haifenglin@berkeley.edu

Gaurav Desai
University of California, Berkeley
gaurav.desai@berkeley.edu

Anish Philip
University of California, Berkeley
anishphilljoe@berkeley.edu

**Abstract**

This paper achieves state of the art results for the ICD code prediction task using the MIMIC-III dataset. This was achieved through the use of Clinical BERT (Alsentzer et al., 2019). embeddings and text augmentation and label balancing to improve F1 scores for both ICD Chapter as well as ICD disease codes. We attribute the improved performance mainly to the use of novel text augmentation to shuffle the order of sentences during training. In comparison to the Top-32 ICD code prediction (Keyang Xu, et. al.) with an F1 score of 0.76, we achieve a final F1 score of 0.75 but on a total of the top 50 ICD codes.

## 1 Introduction

According to studies by Medliminal Healthcare Solutions and Medical Billing Advocates of America, 8 out of 10 bills contain errors. Every year patients in the U.S. pay $68B for these errors (medliminal.com). Medical documentation is the source of generating bills for patients but presents a number of complexities. Medical billers spend substantial time and effort to find relevant text within the documentation and map this text to specific chapters and codes from the ICD classification book. From the text of the documentation, ICD codes are used to classify the type of care that each patient has received and ultimately how much the patient has to pay for their medical treatment.

Our team's goal is to create a medical billing virtual assistant that can improve the efficiency by which medical billing codes are assigned. The medical billing virtual assistant (MBVA) makes use of the MIMIC-III dataset (Johnson, A., Pollard, T., & Mark, R.). From the description from physionet.org: "MIMIC-III is a large, freely-available database comprising de-identified health-related data associated with over forty thousand patients who stayed in critical care units of the Beth Israel Deaconess Medical Center between 2001 and 2012."

The data is a series of related tables that have anonymized patient data. Using these tables, ICD-9 codes are able to be related to clinical note text.

In medicine, because physicians must spend a significant amount of time working on medical billing documentation, this is necessarily time that is not spent with patients. By automating the prediction of medical billing codes, patients will benefit through additional time that physicians can visit with patients.

## 2 Background

The field of natural language processing (NLP) has changed dramatically over the last few years. Embeddings from Language Models (ELMo) was introduced and quickly established itself as a breakthrough (Peters et al., 2018). It utilized Long Short Term Memory (LSTM) networks and was capable of creating word representations that utilized the entire sentence context. ELMo outperformed other models across many different natural language processing (NLP) benchmarks. Shortly thereafter, the Bidirectional Encoder Representations from Transformers (BERT) model was created (Devlin et al., 2018), which represented a further breakthrough in the field of NLP. This model has since become the state of the art for a variety of tasks and is created to take context from a sentence both forward and backwards.

Since that time, the BERT model has been adapted to many different areas of research. In this paper, we examine the performance of Clinical BERT (Alsentzer et al., 2019) to predict ICD codes within the MIMIC-III database. Clinical BERT was trained on the MIMIC-III database, that is, anonymized patient medical notes and patient discharge summaries.

Previous work on MIMIC-III ICD-9 code prediction establishes a number of baselines for F1 scores (Figure 5). In this paper, we aim to use the current state of the art techniques such as Clinical BERT alongside text augmentation and label balancing to achieve improved performance.

## 3 Methods

We focus on a ICD code prediction related to medical billing and make use of the MIMIC-III database.

A variety of model architectures utilizing Clinical BERT were compared and systematically improved.

As our starting point we utilized Clinical BERT with a LSTM as the final layer. We improved this model by moving toward ICD chapter prediction, and by adding a sliding window. By focusing specifically on the most common labels and splitting the model into two, resulted in additional improvement. Next, transfer learning was implemented and the LSTM was swapped out in favor of a CNN. Additional improvement was seen by next balancing the labels through undersampling. Our final model implemented text augmentation to add additional data to the dataset.

There are three main modules in our final model:

1. Preprocessing (Figure 1)
2. ICD Chapter Prediction (Figure 2)
3. ICD Code Prediction (Figure 2)

Preprocessing removes anonymized names and ID tags and replaces medical abbreviations with plain English. As part of preprocessing we focus on identifying Sentence boundaries in the notes that we extract from MIMIC III database. Then we group these sentences into larger chunks and pass them into BERT model to generate embeddings. We standardize the shape of the embeddings by zero padding input text and truncating to 128 tokens per sentence.

Once we have the word embeddings from BERT we pass it through a CNN multi-label classifier to predict the appropriate chapters for the diagnosis codes.

These chapters can also be thought of as specific therapy areas like Respiratory, heart failure, renal insufficiencies, metabolic etc. But in essence each label on this stage points to an actual chapter in the Diagnosis code reference book.

Once we have identified the chapters we move onto more granular disease category predictions. At this stage we have an ensemble of 16 models which leverage the information learned in the previous layer. This improves both speed and precision for training and inference.

In summary, there are a total of 66 labels: 16 labels that predict the diagnosis chapters in layer 1 and 50 labels that predict more granular disease categories in layer 2.

Finally, the trained inference models are deployed within containers using docker-compose such that users can access the model through an API, or a webapp (Figure 3). The details of this architecture is outlined below:

A brief summary of the model architecture:

1. A Flask REST API queries the clinical BERT model and provides the top predicted ICD-9 codes for the given text.

2. AWS Amplify hosts the JS + React front-end, and queries the business logic for the EM-code designations.
3. We make use of the following tools and technologies:
    a. AWS
    b. Docker / Docker-compose
    c. AWS Amplify
    d. React
    e. Flask

The code used in this paper is located here: https://github.com/gauravkdesai/MIDS-W210-Medical_Insurance_Payment_Assistant/

## 4 Results and Discussion

The initial model explored predicted the top 100 ICD-9 codes and achieved a low F1 score of 0.2. Because there are roughly 8,000 frequently referenced ICD codes, high F1 scores are difficult to achieve because of imbalanced labels. ICD is a healthcare classification system that contains codes for disease, symptoms and disorders. The current version, ICD 10 has over 70,000 codes. These ICD codes are split into chapters based on specific therapy areas.

Through substantial effort and experimentation we were able to improve our model performance 4 fold from an F1 score of 0.2 to an F1 score of 0.78 in our final model. The results are detailed in (Figure 4).

The key strategies that led to this improvement are detailed below.

1. Identifying the high level categorization, which, we refer to as 16 chapter labels in layer 1 and 50 disease categories in layer.
2. Moreover we built in transfer learning from layer 1 to layer 2 in order to improve accuracy and reduce training and inference time.
3. We also undersampled the data to achieve balanced labels. This led to an improvement of 15% in F1 score over the original unbalanced dataset.
4. We were able to boost the performance by another 14% in layer 1 and 22% in layer 2 through text augmentation by randomly reordering the sentence grouping, which, was developed in the pre-processing layer

The results from the ensemble of models show improvement upon similar work (Figure 5). We attribute the improved performance mainly to the use of novel text augmentation to shuffle the order of sentences during training. In comparison to the Top-32 ICD code prediction (Keyang Xu, et. al.) with an F1 score of 0.76, we achieve a final F1 score of 0.75 but on a total of the top 50 ICD codes.

## 5 Conclusion

At the time of this paper's publication there are already a variety of different BERT variants that have been trained for specialized tasks. By utilizing the Clinical BERT variant, we demonstrate the ability to predict ICD codes from hospital notes. Using our preprocessing, and two layer model, we are able to go from text notes to billing codes in seconds, while achieving a high F1 score of 0.75 on the top 50 ICD codes.

In the future, as improvements are made to variants of BERT and other models, this will allow for medical coders to concentrate on other tasks and lead to efficiency gains in the healthcare industry. These efficiency gains include but are not limited to the following:

1. The Billing team introduce fewer errors and therefore have higher productivity.
2. Medical Staff are able to spend more time with patients.
3. Insurers have lower costs because of more consistency and less auditing.
4. Patients have additional transparency with their claims.

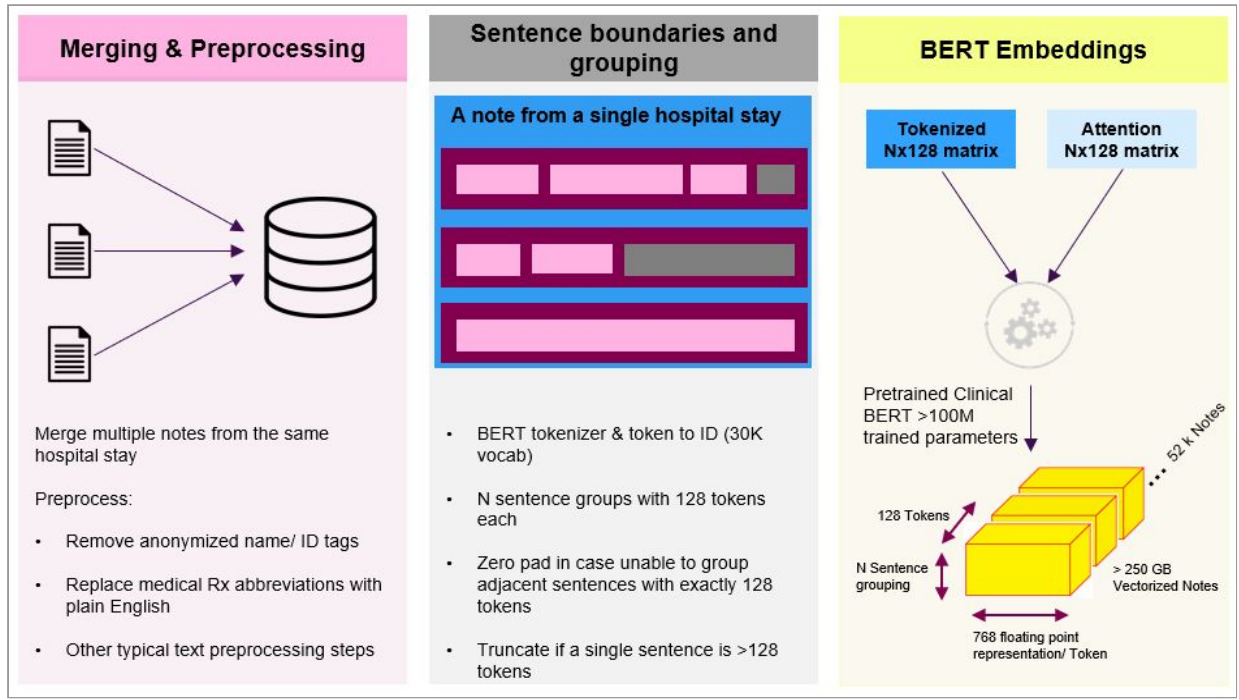

**Figure 1.** Preprocessing

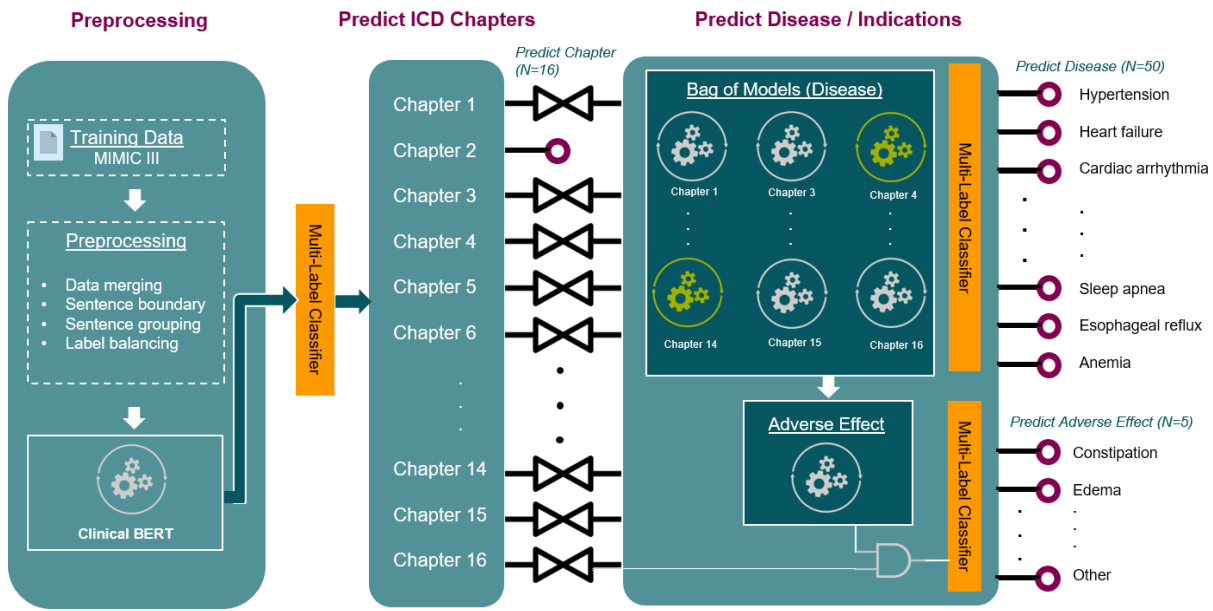

**Figure 2.** Model Architecture with Two Layers

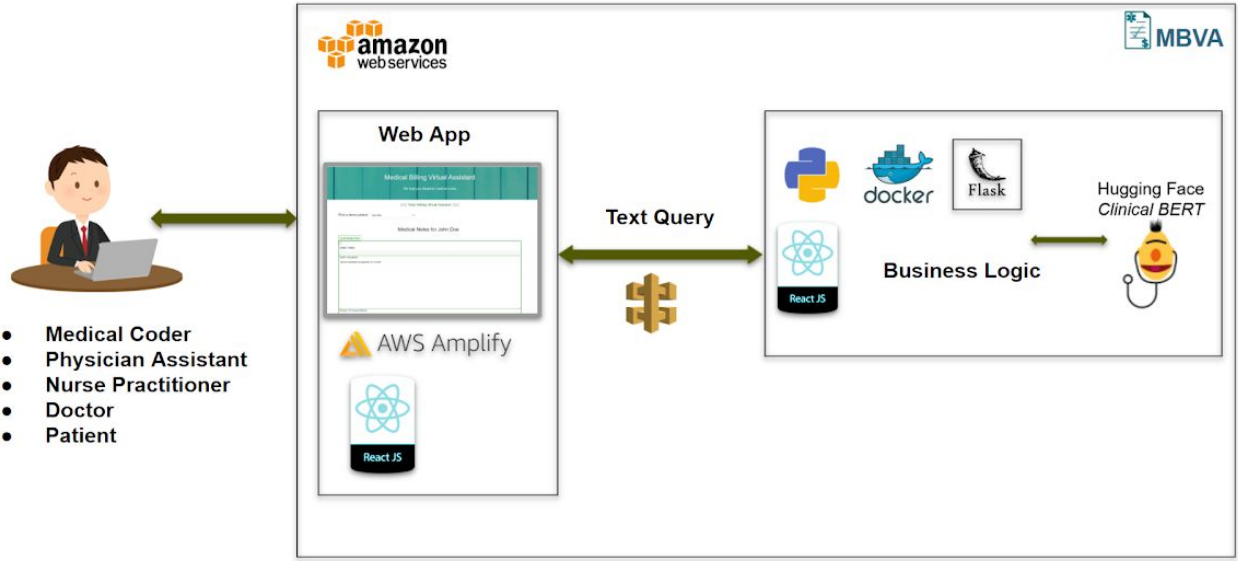

**Figure 3.** Model Deployment Architecture

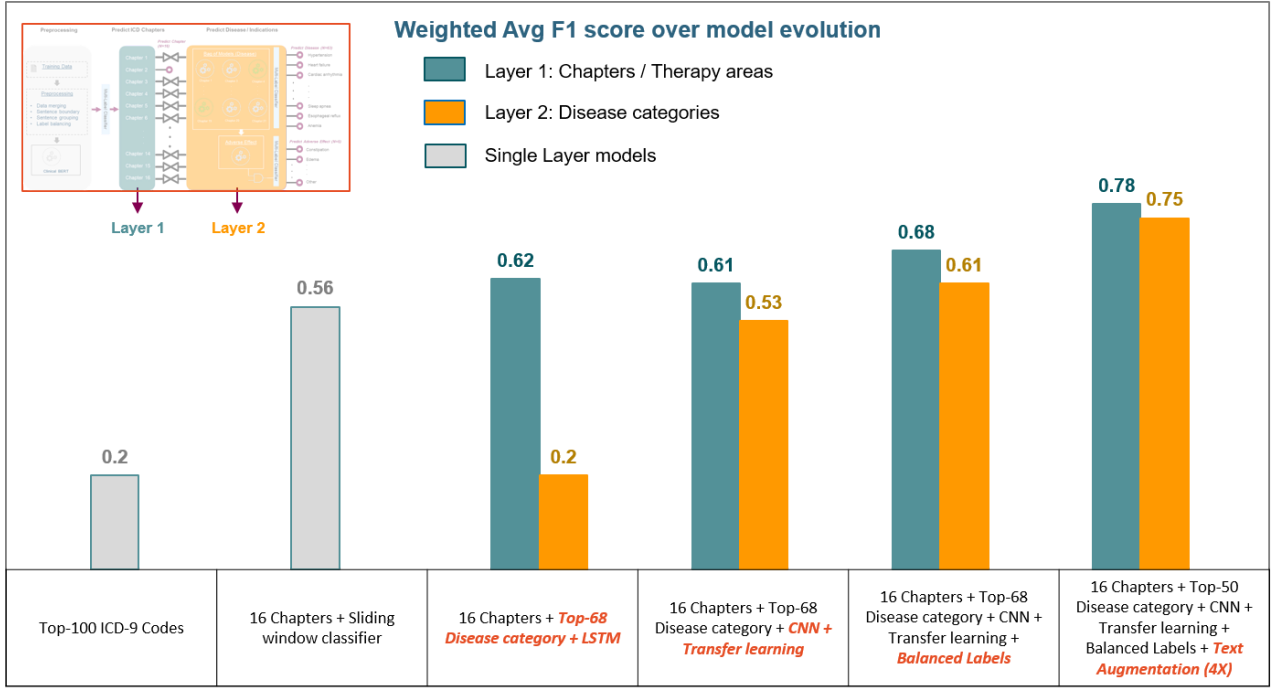

**Figure 4.** Model Improvement Through Time

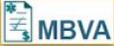

**Figure 5.** Model Comparison to Others